\documentclass[10pt,twocolumn,letterpaper]{article}

\usepackage[pagenumbers]{cvpr} 

\usepackage[dvipsnames]{xcolor}
\newcommand{\red}[1]{{\color{red}#1}}

\usepackage{graphicx}
\usepackage{amsmath}
\usepackage{amssymb}
\usepackage{booktabs}

\definecolor{cvprblue}{rgb}{0.21,0.49,0.74}
\usepackage[pagebackref,breaklinks,colorlinks,citecolor=cvprblue]{hyperref}

\usepackage[capitalize]{cleveref}
\crefname{section}{Sec.}{Secs.}
\Crefname{section}{Section}{Sections}
\Crefname{table}{Table}{Tables}
\crefname{table}{Tab.}{Tabs.}

\def\confName{CVPR}
\def\confYear{2024}

\usepackage{color}
\usepackage{xcolor}
\definecolor{deepGreen}{RGB}{0,153,0}
\definecolor{orange}{RGB}{255,125,0}
\def\red#1{\textcolor[rgb]{1,0,0}{#1}}

\def\graytext#1{\textcolor[RGB]{85,85,85}{#1}}

\definecolor{sainone}{RGB}{236, 242, 249}
\definecolor{saintwo}{RGB}{255, 230, 204}

\newcommand{\keypoint}[1]{\vspace{0.1cm}\noindent\textbf{#1}\;}
\newcommand{\cut}[1]{}
\usepackage{multirow}
\usepackage{amsmath}
\usepackage{amssymb}
\usepackage{mathtools}
\usepackage{arydshln}
\usepackage{soul}
\usepackage{nicefrac}
\usepackage{booktabs}
\usepackage{stmaryrd}
\usepackage{caption}
\usepackage{graphicx}
\usepackage{colortbl}
\definecolor{gray}{gray}{0.9}
\definecolor{pink}{RGB}{255, 234, 232}
\sethlcolor{pink}
\usepackage[accsupp]{axessibility}
\usepackage{scalerel,graphicx,xparse}

\newcommand{\MYhref}[3][blue]{\href{#2}{\color{#1}{#3}}}

\makeatletter
\apptocmd\@maketitle{{\myfigure{}\par}}{}{}
\makeatother
\usepackage{capt-of,etoolbox}

\makeatletter
\newcommand\notsotiny{\@setfontsize\notsotiny\@vipt\@viipt}
\makeatother

\usepackage{xcolor,pifont}
\newcommand*\colourcheck[1]{%
  \expandafter\newcommand\csname #1check\endcsname{\textcolor{#1}{\ding{52}}}%
}
\newcommand*\colourcross[1]{%
  \expandafter\newcommand\csname #1cross\endcsname{\textcolor{#1}{\ding{55}}}%
}
\colourcheck{blue}
\colourcheck{green}
\colourcheck{black}
\colourcross{red}
\colourcross{black}

\title{\vspace{-0.8cm}Text-to-Image Diffusion Models are Great Sketch-Photo Matchmakers \vspace{-0.7cm}}

\author{\MYhref[cvprblue]{https://subhadeepkoley.github.io}{Subhadeep Koley}\textsuperscript{1,2} \hspace{.2cm} \MYhref[cvprblue]{https://ayankumarbhunia.github.io}{Ayan Kumar Bhunia}\textsuperscript{1} \hspace{.2cm} \MYhref[cvprblue]{https://aneeshan95.github.io}{Aneeshan Sain}\textsuperscript{1} \hspace{.2cm}  \MYhref[cvprblue]{https://www.pinakinathc.me}{Pinaki Nath Chowdhury}\textsuperscript{1} \\ \MYhref[cvprblue]{https://www.surrey.ac.uk/people/tao-xiang}{Tao Xiang}\textsuperscript{1,2} \hspace{.2cm} \MYhref[cvprblue]{https://www.surrey.ac.uk/people/yi-zhe-song}{Yi-Zhe Song}\textsuperscript{1,2} \\
\textsuperscript{1}SketchX, CVSSP, University of Surrey, United Kingdom.  \\
\textsuperscript{2}iFlyTek-Surrey Joint Research Centre on Artificial Intelligence.\\
{\tt\small \{s.koley, a.bhunia, a.sain, p.chowdhury, t.xiang, y.song\}@surrey.ac.uk}\\
\small \url{https://subhadeepkoley.github.io/DiffusionZSSBIR}
}
\vspace{-0.5cm}

\newcommand\myfigure{
\centering
\vspace{-0.9cm}
\captionsetup{type=figure} 
    \includegraphics[width=\textwidth]{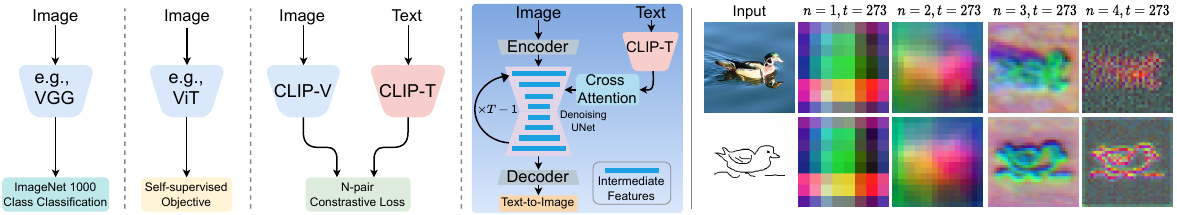}
    \vspace{-0.7cm}
\captionof{figure}{Sketch-based image retrieval frameworks~\cite{yu2016sketch, yelamarthi2018zero, collomosse2019query} usually employ ImageNet pre-trained CNNs~\cite{bhunia2020sketch, dey2019doodle, yu2016sketch}, JFT-trained vision transformers (ViT)~\cite{sain2023exploiting, ribeiro2020sketchformer}, or visual encoders of vision-language models like CLIP~\cite{sain2023clip} as \textit{backbone feature extractors}. Rich knowledge from large-scale pre-training offers a good initialisation, which when further fine-tuned on sketch-photo datasets, performs way better than training from random initialisation \cite{pang2020solving}. While one can extract features either by discarding the classification head for ImageNet pre-trained models, auxiliary task head for self-supervised models, or by using CLIP's visual encoder, text-to-image diffusion models (\eg, stable diffusion) lack any specific feature embedding space. However, we find that its intermediate representations implicitly hold robust cross-modal features at multiple granularities. Unlike prior SBIR backbones, pre-trained with \textit{discriminative tasks}, we propose to leverage denoising diffusion models pre-trained with text-to-image \textit{generative tasks} to bridge the sketch-photo domain gap. Being a text-to-image generation model trained on a large corpus of text-image pairs (LAION~\cite{schuhmann2021laion, schuhmann2022laion}), it holds both semantic and shape prior~\cite{tang2023emergent}. PCA representation~\cite{tang2023emergent} (right) of intermediate UNet features (sketch/photo) from different upsampling blocks (details in \cref{sec:pilot_1}) depict that they share significant semantic similarity (denoted by similar colours).}
\label{fig:teaser}
\vspace{+0.3cm}
}

\begin{document}
\maketitle

\begin{abstract}
\vspace{-0.2cm}

This paper, for the first time, explores text-to-image diffusion models for Zero-Shot Sketch-based Image Retrieval (ZS-SBIR). We highlight a pivotal discovery: the capacity of text-to-image diffusion models to seamlessly bridge the gap between sketches and photos. This proficiency is underpinned by their robust cross-modal capabilities and shape bias, findings that are substantiated through our pilot studies. In order to harness pre-trained diffusion models effectively, we introduce a straightforward yet powerful strategy focused on two key aspects: selecting optimal feature layers and utilising visual and textual prompts. For the former, we identify which layers are most enriched with information and are best suited for the specific retrieval requirements (category-level or fine-grained). Then we employ visual and textual prompts to guide the model's feature extraction process, enabling it to generate more discriminative and contextually relevant cross-modal representations. Extensive experiments on several benchmark datasets validate significant performance improvements.

\end{abstract}

\vspace{-0.6cm}
\section{Introduction}
\vspace{-0.1cm}
Diffusion models~\cite{ho2020denoising, ho2022classifier, rombach2022high, dhariwal2021diffusion}, celebrated for their proficiency in generating photorealistic images, have witnessed remarkable advances in the field of computer vision. Ranging from image generation~\cite{rombach2022high, dhariwal2021diffusion, ramesh2021zero, saharia2022photorealistic} to semantic segmentation~\cite{xu2023open, baranchuk2021label} and image-to-image translation~\cite{tumanyan2023plug, mou2023t2i, zhang2023adding}, these models have ignited transformations across various domains within the computer vision discipline~\cite{tang2023emergent, li2023your, de2023medical, chen2023democaricature}. In this paper, we embark on an uncharted journey, delving into the intrinsic feature representations of diffusion models, with a primary emphasis on Zero-Shot Sketch-based Image Retrieval (ZS-SBIR).

At the heart of our exploration lies the assertion that text-to-image diffusion models~\cite{rombach2022high, dhariwal2021diffusion} excel as ``matchmakers'', seamlessly connecting the realms of sketches and photos. We substantiate this claim through a series of pilot studies, unveiling two fundamental properties inherent in diffusion models, which are pivotal for the success of ZS-SBIR~\cite{yelamarthi2018zero, sain2023clip}. First and foremost, these models naturally transcend modalities, imbued with the ability to bridge the gap between different types of data, providing a robust foundation for cross-modal retrieval. Furthermore, we unearth a pronounced \textit{shape bias} nestled within their joint embeddings, rendering them exquisitely well-suited for the intricacies of SBIR. It is worth noting that zero-shot capabilities virtually come as an added bonus, thanks to the open-vocab nature of large-scale pre-trained diffusion models~\cite{rombach2022high, dhariwal2021diffusion} (\eg, Stable Diffusion~\cite{rombach2022high}).

Nonetheless, effectively harnessing pre-trained diffusion models~\cite{rombach2022high} for ZS-SBIR may initially seem not entirely straightforward. These models encapsulate a diverse array of features across different layers and time-steps~\cite{rombach2022high}, requiring careful selection to align with the intricacies of the specific task at hand. Moreover, their pre-training with a text-to-image generation objective~\cite{rombach2022high} endows them with a heavy bias to generate \textit{text-influenced image features}, moving them away from the conventional sketch-photo visual matching~\cite{sain2023clip} problem encountered in SBIR.

Our solution to this seemingly intricate problem, however, is elegantly simple. It revolves around two key principles: \textit{(i)} identifying the layers and time-steps most enriched with information for SBIR, and \textit{(ii)} optimising the interaction with Stable Diffusion (SD) to effectively condition feature extraction. In addressing the former, we shed light on the optimal layers and time-steps for feature extraction, tailored to various ZS-SBIR tasks, including its fine-grained variant. Our findings suggest that, for category-level retrieval (\ie, ZS-SBIR), {features from the first few UNet upsampling blocks} prove most effective, while for fine-grained retrieval (\ie, ZS-FG-SBIR), latter {upsampling blocks} excel. To adapt the pre-trained Stable Diffusion model~\cite{rombach2022high} for SBIR, we employ task-specific visual prompts within both sketch and photo branches. The SD model~\cite{rombach2022high}, trained with a text-to-image synthesis objective, boasts rich visio-lingual joint information~\cite{rombach2022high}. To leverage this text-enriched knowledge to the fullest, we introduce textual prompt learning. We acquire both visual and textual prompts using a simple triplet-based objective, all the while maintaining the integrity of the frozen SD model~\cite{rombach2022high}. During inference, we apply the learned visual prompt to the query sketch and pass it through the frozen SD model~\cite{rombach2022high}, conditioned on the acquired textual prompt. This process yields the query feature, which is subsequently compared with pre-computed gallery features, facilitating retrieval.

In summary, \textit{(i)} we unveil the latent potential of diffusion models as backbone {feature extractors} for ZS-SBIR, substantiated by empirical evidence and comprehensive analysis. \textit{(ii)} we introduce innovative design strategies, including soft prompt learning and visual prompting, designed to address the challenges of \textit{all} forms of zero-shot SBIR tasks. \textit{(iii)} through extensive experimentation on standard datasets, we demonstrate marked improvements in performance compared to conventional ZS-SBIR approaches.

\section{Related Works}
\vspace{-0.2cm}
\keypoint{Sketch-Based Image Retrieval (SBIR).}
SBIR began as a category-level task aimed at retrieving a photo of the same category as the query sketch. Evolving from classical approaches~\cite{saavedra2014sketch, saavedra2015sketch, hu2013performance, qi2015making}, deep-learning ones \cite{liu2017deep, yang2020deep, koley2024how, collomosse2019query} usually trained Siamese-like networks \cite{collomosse2019query}, typically based on CNNs \cite{dey2019doodle}, RNNs \cite{xu2018sketchmate} or Transformers \cite{ribeiro2020sketchformer} to fetch similar photos over a distance-metric in a cross-modal joint embedding space~\cite{collomosse2017sketching}. It was enhanced further by unsupervised multi-clustering based re-ranking~\cite{wang2019sketch}, learning discriminative structure representation via dynamic landmark discovery \cite{zhang2019learning}, incremental learning~\cite{bhunia2022doodle}, etc.

\keypoint{Fine-Grained SBIR.}
Given a sketch, FG-SBIR aims to fetch {its} \textit{target}-photo from a gallery of same-category photos. Since its inception~\cite{li2014fine}, numerous works have emerged \cite{yu2016sketch, song2017deep, pang2019generalising, bhunia2021more, sain2020cross}, aided by new datasets having fine-grained sketch-photo association \cite{yu2016sketch,sangkloy2016sketchy, chowdhury2022fs}. From a deep-triplet ranking based Siamese network~\cite{yu2016sketch} learning a joint sketch-photo manifold, FG-SBIR was enhanced with attention mechanism using higher-order~\cite{song2017deep} losses, hybrid cross-domain image generation \cite{pang2017cross}, textual tags \cite{song2017fine}, local feature alignment \cite{xu2021dla} and mixed-modal jigsaw solving based pre-training tasks \cite{pang2020solving}, to name a few. While some addressed inherent sketch-traits, like hierarchy \cite{sain2020cross}, style-diversity \cite{sain2021stylemeup}, or redundancy of strokes \cite{bhunia2022sketching} others explored applicability like early-retrieval \cite{bhunia2020sketch}, cross-category generalisation \cite{pang2019generalising, bhunia2022adaptive}, and overcoming data-scarcity \cite{sain2023exploiting, bhunia2021more}. Recently FG-SBIR was extended to scene-level (retrieve a scene image, given a scene-sketch) via cross-modal region-associativity \cite{chowdhury2022partially}, or with an optional text-query \cite{chowdhury2023scenetrilogy}.

\keypoint{Zero-Shot SBIR.} Towards tackling data-scarcity, ZS-SBIR aims at generalising knowledge learned from \textit{seen} training classes to \textit{unseen} (disjoint) testing categories. Introduced by Yelamarthi \etal~\cite{yelamarthi2018zero} where photo-features were approximated from sketches via VAE-based~\cite{kingma2013auto} image-to-image translation, to generalise onto unseen classes,  later works shifted to exploiting word2vec~\cite{mikolov2013distributed}  representation of class labels~\cite{dey2019doodle, dutta2019semantically} for semantic transfer. While \cite{dutta2019semantically} attempted to align sketch, photo and semantic representations via adversarial training, others aimed to minimise the sketch-photo domain gap using a gradient reversal layer \cite{dey2019doodle}, a conditional-VAE based graph convolution network \cite{zhang2020zero}, a shared ViT~\cite{tian2022tvt} backbone, employing prototype-based~\cite{wang2022prototype} selective knowledge distillation, and a test-time training paradigm \cite{sain2022sketch3t} to name a few. While semantic transfer in all such works was mostly limited to using word embeddings directly~\cite{dutta2019semantically,zhang2020zero,wang2021transferable} or indirectly~\cite{tian2021relationship,liu2019semantic,wang2022prototype}, very recently Sain \etal \cite{sain2023clip} adapted CLIP \cite{radford2021learning}  to exploit its high generalisability for semantic transfer. Instead of using feature extractors trained on \textit{discriminative} tasks, we propose using pre-trained text-to-image \textit{generative} backbone for all kinds of SBIR tasks.

\keypoint{Backbones for SBIR.} State-of-the-art backbones for SBIR can be broadly categorised as – \textit{(a) Standard pre-trained}~\cite{chowdhury2022partially, bhunia2022sketching, bhunia2021more}, \textit{(b) VAE-based formulations}~\cite{sain2021stylemeup, zhang2020zero, yelamarthi2018zero}, \textit{(c) Self-supervised pre-trained}~\cite{pang2020solving, li2023photo}, and \textit{(d) Foundational models}~\cite{sain2023clip, sangkloy2022sketch}. \textit{Standard} models typically use ImageNet pre-trained CNNs~\cite{chowdhury2022partially, bhunia2022sketching, bhunia2021more} with specific design choices like – spatial attention~\cite{song2017deep}, conditional stroke recovery~\cite{ling2022conditional}, cross-modal pairwise-embedding~\cite{sain2020cross}, cross-attention~\cite{lin2023zero}, etc. While, \textit{VAE-based}~\cite{kingma2013auto} models were used in sketch-photo translation to minimise sketch-photo domain gap~\cite{yelamarthi2018zero}, meta-learning content-style disentanglement~\cite{sain2021stylemeup} etc., other works focused on \textit{self-supervised pre-training} either via leveraging the neighbourhood topology induced by large-scale photo pre-training~\cite{li2023photo} or by solving mixed-modal jigsaw as a pretext task~\cite{pang2020solving}. Finally, the recent rise of \textit{foundational models}~\cite{rombach2022high, radford2021learning, li2022blip}, nudged several works to adapt them as backbones, by either training CLIP~\cite{radford2021learning} end-to-end~\cite{sangkloy2022sketch} for standard SBIR or fine-tuning the $\mathtt{LayerNorm}$ layers~\cite{sain2023clip} of a pre-trained CLIP model with prompt learning for ZS-SBIR. In this paper, we aim to address \textit{all} aspects (\ie, category-level and fine-grained) of zero-shot SBIR, utilising the inherent zero-shot potential~\cite{li2023your} of a pre-trained Stable Diffusion Model~\cite{rombach2022high}.

\keypoint{Diffusion Models for Vision Task.} Recently, the diffusion model~\cite{ho2020denoising} has become the de facto standard for image generation~\cite{dhariwal2021diffusion} and editing~\cite{ruiz2023dreambooth}. Essentially, it converts an image $\mathbf{x}_0$ into a noisy image $\mathbf{x}_T$ by iteratively adding random Gaussian noise to $\mathbf{x}_0$ in $T$ time-steps~\cite{ho2020denoising}. The reverse, recovers $\mathbf{x}_0$ from $\mathbf{x}_T$ over $T$ denoising steps~\cite{ho2020denoising}. Subsequent developments of classifier-free guidance~\cite{ho2022classifier} and latest latent diffusion model~\cite{rombach2022high} lead to numerous controllable image generation frameworks like DALL-E~\cite{ramesh2021zero}, Make-A-Scene~\cite{gafni2022make}, Imagen~\cite{saharia2022photorealistic}, T2I-Adapter~\cite{mou2023t2i}, ControlNet~\cite{zhang2023adding}, CogView~\cite{ding2021cogview}, etc. This gave rise to various works \cite{ruiz2023dreambooth,kawar2023imagic,hertz2022prompt,meng2021sdedit,mikaeili2023sked,gal2022image} on realistic image-edits leveraging frozen diffusion models. Additionally, \textit{diffusion} has found use in several vision tasks like classification~\cite{li2023your}, semantic~\cite{baranchuk2021label} and panoptic segmentation~\cite{xu2023open}, image-to-image translation~\cite{tumanyan2023plug, koley2023its}, medical imaging~\cite{de2023medical}, image correspondence~\cite{tang2023emergent}, object detection~\cite{xu20233difftection}, etc. Despite its advances, the efficacy of pre-trained diffusion models for cross-modal zero-shot retrieval remains under-explored. Here, we thus aim to explore the inherent zero-shot cross-modal retrieval potential of a frozen stable diffusion model.

\vspace{-0.2cm}
\section{Revisiting Text-to-Image Diffusion Models}
\vspace{-0.2cm}
\label{sec:diffusion_background}
\keypoint{Overview.} Diffusion models generate images by progressive removal of noise from an initial pure 2D Gaussian noise~\cite{ho2020denoising, dhariwal2021diffusion}. It relies on two complementary random processes, namely, \textit{``forward''} and \textit{``reverse''} diffusion~\cite{ho2020denoising}. The \textit{forward} process iteratively adds Gaussian noise of varying magnitudes to a clean image $\mathbf{x}_0 \in \mathbb{R}^{h\times w\times 3}$  from the training dataset, for {$t$} time-steps to produce a noisy image $\mathbf{x}_t = \sqrt{\Bar{\alpha}_t}\mathbf{x}_{0} + \sqrt{1-\Bar{\alpha}_t}\epsilon$. Here, $\epsilon$$\sim$$\mathcal{N}(0,\mathbf{I})$ is the added noise, $\{\alpha_t\}_1^T$ is a pre-defined noise schedule where $\Bar{\alpha}_t=\prod_{k=1}^t \alpha_k$ with $t$$\sim$$U(0,T)$. {When $T$ is large enough, the resulting image ($\mathbf{x}_T \in \mathbb{R}^{h\times w\times 3}$) approximates pure noise~\cite{ho2020denoising}.} The \textit{reverse} process involves training a denoising UNet~\cite{ronneberger2015u} $\mathcal{F}_\theta$, that tries to estimate the input noise $\epsilon \approx \mathcal{F}_\theta(\mathbf{x}_t,t)$ from the noisy image $\mathbf{x}_t$ at every time-step $t$. Once trained with an $l_2$ objective~\cite{ho2020denoising}, $\mathcal{F}_\theta$ can reverse the effect of the forward diffusion. During inference, starting from a random pure $2D$ noise $\mathbf{x}_T$ sampled from a Gaussian distribution, $\mathcal{F}_\theta$ is applied iteratively (for $T$ time-steps) to estimate and subtract the noise from each time-step to get a cleaner image $\mathbf{x}_{t-1}$, eventually leading to one of the cleanest images $\mathbf{x}_0$ from the original target distribution~\cite{ho2020denoising}.

\keypoint{Text-Conditioned Diffusion Model.} This unconditional process could be made ``conditional’’ by influencing the $\mathcal{F}_\theta$ with auxiliary guiding signals $\mathbf{p}$ (\eg, class labels~\cite{ho2022cascaded}, textual prompts~\cite{nichol2021glide, rombach2022high, kim2022diffusionclip, ramesh2022hierarchical, saharia2022photorealistic}, dense semantic maps~\cite{zhang2023adding, mou2023t2i}, etc.). A pre-trained CLIP text encoder~\cite{radford2021learning} $\mathcal{T}(\cdot)$ converts textual prompt $\mathbf{p}$ (textual description) into tokenised embedding $\mathbf{T}_p=\mathcal{T}(\mathbf{p})\in\mathbb{R}^{77\times d}$ that influences the denoising process via cross-attention to generate spatial attention maps for each word token~\cite{rombach2022high}. Spatial attention maps at each time-step $t$ control the visio-linguistic interaction between spatial feature maps and text tokens. Here, we use the text-conditional Latent Diffusion Model (SD)~\cite{rombach2022high}, where the diffusion process occurs on the latent space instead of the pixel space, for faster training~\cite{rombach2022high}.

\keypoint{Stable Diffusion Architecture.} Given an image and text-prompt pair ($\mathbf{x,p}$), SD~\cite{rombach2022high} first uses the encoder $\mathcal{E}(\cdot)$ from a pre-trained variational autoencoder -- VAE (consists of an encoder $\mathcal{E}(\cdot)$ and decoder $\mathcal{D}(\cdot)$ in cascade)~\cite{kingma2013auto} to convert the input image $\mathbf{x}_0 \in \mathbb{R}^{h\times w\times 3}$ to a latent image $\mathbf{z}_0 =\mathcal{E}(\mathbf{x}_0) \in \mathbb{R}^{\frac{h}{8}\times \frac{w}{8}\times d}$. Later, a \textit{time-conditional denoising UNet}~\cite{ronneberger2015u} $\mathcal{F}_\theta(\cdot)$ is trained to denoise directly on the latent space. SD's UNet architecture consists of $12$ encoding layers, $1$ middle layer, and $12$ skip-connected decoding layers~\cite{rombach2022high}. Among the $12$ encoding and $12$ decoding layers, there are $4$ downsampling $\{\mathcal{F}_{\mathbf{d}}^1,\mathcal{F}_{\mathbf{d}}^2,\mathcal{F}_{\mathbf{d}}^3,\mathcal{F}_{\mathbf{d}}^4\}$ and $4$ upsampling $\{\mathcal{F}_{\mathbf{u}}^1,\mathcal{F}_{\mathbf{u}}^2,\mathcal{F}_{\mathbf{u}}^3,\mathcal{F}_{\mathbf{u}}^4\}$ layers respectively. $\mathcal{F}_{\theta}(\cdot)$ takes three inputs -- \textit{(i)} the $t^\text{th}$ step noisy latent $\mathbf{z}_t = \sqrt{\Bar{\alpha}_t}\mathbf{z}_{0} + \sqrt{1-\Bar{\alpha}_t}\epsilon$, \textit{(ii)} the scalar time-step value $t$, and \textit{(iii)} textual embedding $\mathbf{T}_p = \mathcal{T}(\mathbf{p})$ $\in$ $\mathbb{R}^{77\times d}$. Both time-embedding and textual embedding influence the intermediate feature maps of the UNet via cross-attention~\cite{rombach2022high}. $\mathcal{F}_\theta(\cdot)$ then predicts the noise map $\Hat{\epsilon}_{t}$ (same size as $\mathbf{z}_t$) corresponding to that time-step. Mathematically,

\vspace{-0.35cm}
\begin{equation}
    \mathcal{F}_\theta : (\mathbf{z}_t, t, \mathbf{T}_p) \rightarrow \Hat{\epsilon}_{t}
    \vspace{-0.1cm}
\end{equation}

Parameters of $\mathcal{F}_\theta$ are learned over $l_2$ loss as: $\mathcal{L}_\text{SD}=\mathbb{E}_{\mathbf{z}_t,t,\mathbf{T}_p,\epsilon} ({||\epsilon-\mathcal{F}_{\theta}(\mathbf{z}_t,t,\mathbf{T}_p)||}_2^2)$. During text-to-image inference, SD \textit{discards} $\mathcal{E}(\cdot)$, directly sampling a noisy latent image $\mathbf{z}_T$ from Gaussian distribution~\cite{rombach2022high}. It then  estimates noise from $\mathbf{z}_T$ iteratively via $\mathcal{F}_\theta$ (conditioned on $\mathbf{p}$) to obtain a clean latent $\Hat{\mathbf{z}}_0\in \mathbb{R}^{\frac{h}{8}\times \frac{w}{8}\times d}$ after $T$ iterations. Passing $\Hat{\mathbf{z}}_0$ via frozen VAE decoder, generates the final image $\Hat{\mathbf{x}}=\mathcal{D}(\Hat{\mathbf{z}}_0)\in \mathbb{R}^{h\times w\times 3}$~\cite{rombach2022high}.

\vspace{-0.1cm}
\begin{figure}[!htbp]
    \centering
    \includegraphics[width=1\linewidth]{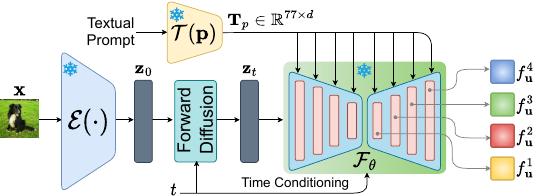}
    \vspace{-0.6cm}
    \caption{Feature extraction via text-to-image diffusion model.}
    \label{fig:feat_ext}
    \vspace{-0.6cm}
\end{figure}

\section{Feature Extraction via Stable Diffusion}
\vspace{-0.1cm}
\label{sec:feat_ext}

Intermediate feature maps of $\mathcal{F}_\theta$ holding rich semantic information have been used in several downstream vision tasks like segmentation~\cite{xu2023open}, classification~\cite{li2023your}, correspondence-learning~\cite{hedlin2023unsupervised}, etc. The extracted feature depends on two \textit{task-specific} design choices -- \textit{(i)} the time-step $t$, and \textit{(ii)} the selection of a particular intermediate feature map at that time-step. Here (\cref{fig:feat_ext}), we aim to use these information-rich interim representations as backbone features for ZS-SBIR. In particular, given an image and text-prompt pair ($\mathbf{x},\mathbf{p}$), and a time-step $t$, we first generate the latent image $\mathbf{z}_0$$=$$\mathcal{E}(\mathbf{x})$. We then add noise from time-step $t$ to transform $\mathbf{z}_0$ to its $t^\text{th}$-step noisy latent image $\mathbf{z}_t$ (forward diffusion). Now we feed,  -- \textit{(i)} the noisy latent $\mathbf{z}_t$, \textit{(ii)} scalar time-step value $t$, and \textit{(iii)} textual embedding $\mathbf{T}_p=\mathcal{T}(\mathbf{p})$ into $\mathcal{F}_\theta(\cdot)$ to extract corresponding intermediate features from upsampling layers $\{\mathcal{F}_{\mathbf{u}}^1,\mathcal{F}_{\mathbf{u}}^2,\mathcal{F}_{\mathbf{u}}^3,\mathcal{F}_{\mathbf{u}}^4\}$. For example, an image of size $\mathbb{R}^{256\times 256\times 3}$ would yield feature maps of $f_{\mathbf{u}}^1$$=$$\mathbb{R}^{8\times 8\times 1280}$, $f_{\mathbf{u}}^2$$=$$\mathbb{R}^{16\times 16\times 1280}$, $f_{\mathbf{u}}^3$$=$$\mathbb{R}^{32\times 32\times 640}$, and $f_{\mathbf{u}}^4$$=$$\mathbb{R}^{32\times 32\times 320}$ from $\{\mathcal{F}_{\mathbf{u}}^1,\mathcal{F}_{\mathbf{u}}^2,\mathcal{F}_{\mathbf{u}}^3,\mathcal{F}_{\mathbf{u}}^4\}$ respectively. We then perform global max-pooling on these feature maps ($f_{\mathbf{u}}^n$) to produce $d$-dimensional feature vectors ($\Hat{f_{\mathbf{u}}^n}$). In the absence of $\mathbf{p}$, the embedding for a \textit{null prompt}~\cite{tang2023emergent} is used.

\vspace{-0.2cm}
\section{Pilot Study: Analysis and Insights}
\vspace{-0.2cm}
\label{sec:pilot_1}
\keypoint{Cross-modal Correspondence.}
High-quality image generation~\cite{rombach2022high, ruiz2023dreambooth} from SD models inspired us to further investigate their internal representations and explore whether a \textit{frozen} SD model can deliver cross-modal shape correspondence~\cite{tang2023emergent}. Accordingly, we hypothesise that -- \textit{(i)} although aimed at image generation~\cite{rombach2022high}, the UNet denoiser's internal feature maps embed highly localised hierarchical semantic information; \textit{(ii)} this localised semantic knowledge is robust across various image modalities (\eg, image \vs sketch). To validate, we extracted SD internal features of a few photos and their corresponding sketches (\cref{sec:feat_ext}) at $t$$=$$273$ (ablated in \cref{sec:abal}) with a fixed prompt $\mathtt{``a~photo~of~[CLASS]"}$~\cite{tang2023emergent}. For visualising the high-dimensional features, we perform PCA on feature maps $f_{\mathbf{u}}^n;n\in\{1,2,3,4\}$, and render the first three principal components~\cite{tang2023emergent, luo2023diffusion} as RGB images (\cref{fig:teaser} right). Feature representations from different upsampling layers here depict the different feature hierarchies. While earlier layers ($n$$=$$1,2$) delineate low-frequency coarse-level features (\eg, shape and structure), later ones ($n$$=$$3,4$) depict high-frequency fine-grained features (\eg, image gradients), thus validating our first hypothesis. We can also see how sketch and photo features from different decoder levels share significant \textit{semantic similarity} (shown in \textit{same} colour) albeit originating from different modalities that hold significant visual disparity (\eg, sparse-binary sketch \vs pixel-perfect photo)~\cite{sain2021stylemeup} -- supporting our second presumption. This motivates us to leverage the frozen SD model's internal features for both category-level and fine-grained ZS-SBIR.

\begin{figure}[!htbp]
    \centering
    \includegraphics[width=\columnwidth]{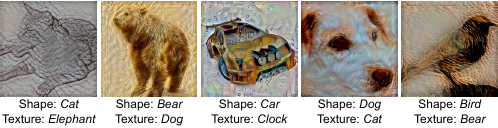}
    \vspace{-0.7cm}
    \caption{Texture-altered images from cue-conflict~\cite{geirhos2018imagenet} dataset.}
    \label{fig:texture-bias}
    \vspace{-0.4cm}
\end{figure}

\keypoint{Reduced Texture-bias.} ImageNet \cite{deng2009imagenet} pre-trained discriminative CNN \cite{simonyan2015very} backbones are highly biased towards texture features \cite{geirhos2018imagenet, jaini2023intriguing}. Now, \textit{sketches} being binary and sparse line drawings, do not provide any colour/texture cues in context of image retrieval \cite{koley2023you, koley2023picture}. Thus, SBIR being a \textit{shape-matching} \cite{bhunia2022adaptive, bhunia2021more} problem should ideally be biased towards \textit{shape} instead of texture features. We posit that the texture-bias of discriminative CNN backbones largely bottlenecks existing SBIR performance. To validate this, we use the cue-conflict dataset \cite{geirhos2018imagenet} containing images across $16$ classes, where style-transfer \cite{gatys2016image} is applied to alter the texture (style) of the images with random textures of images from other classes, keeping the shape (content) same (\cref{fig:texture-bias}). Out of them, we select $10$ classes overlapping with Sketchy~\cite{sangkloy2016the} to form our retrieval gallery. We train a frozen VGG-16 \cite{simonyan2015very} backbone followed by \emph{one} learnable FC-layer with triplet loss on real sketch-photo pairs from Sketchy \cite{sangkloy2016the}. We observe that the SBIR accuracy (tested on cue-conflict gallery) of VGG-16 \textit{discriminative} backbone is much lower ($49.14\%$) than our method using the SD \textit{generative} backbone ($72.82\%$). Furthermore, asking a group of human observers to categorise these texture-altered images, resulted in high error-consistency ($90.16\%$) \cite{jaini2023intriguing} with our method. This shows generative backbones to be biased towards \textit{shape} and handle \textit{adversarial} examples \cite{jaini2023intriguing} better, like the human visual system, tailoring them for ZS-SBIR.

\vspace{-0.2cm}

\vspace{-0.2cm}
\section{Background: Baseline Zero-Shot SBIR}
\vspace{-0.2cm}
\keypoint{Category-Level ZS-SBIR.} For a query sketch $s$ of any class, this aims to fetch a photo $p_i^j$ of the \textit{same} class, from a gallery $\mathcal{G}$$=$$\{p_1^j,\cdots, p_{N_j}^j\}_{j=1}^{N_c}$ containing photos from $N_c$ classes with $N_j$ photos per class. Usually, an encoder $\mathcal{B}(\cdot)$ (separate weights for sketch and photo branch), is trained to extract $d$-dimensional feature $f_{i} = \mathcal{B}(\mathcal{I}):\mathbb{R}^{h\times w\times 3}\rightarrow \mathbb{R}^{d}$, over a triplet loss~\cite{yu2016sketch} that aims to minimise the distance $\delta(\cdot,\cdot)$ between features of an anchor sketch ($f_s$) and its matching photo ($f_p$) from the same class, while increasing that from a non-matching photo ($n$) feature $f_n$ of a \textit{different} class~\cite{yu2016sketch}. With margin $\mu>0$, the triplet loss becomes:

\vspace{-0.3cm}
\begin{equation}
\mathcal{L}_\text{trip} = \mathtt{max}\{0,\mu+\delta(f_s,f_p)-\delta(f_s,f_n)\}
\label{equ:basic_triplet}
\vspace{-0.2cm}
\end{equation}

While conventional SBIR evaluates on classes \textit{\underline{s}een} during training, $\mathcal{C}^S=\{c_1^S, \cdots, c_{N}^S\}$, ZS paradigm~\cite{yelamarthi2018zero} evaluates on those \textit{\underline{u}nseen}, $\mathcal{C}^U=\{c_1^U, \cdots, c_{M}^U\}$, $\therefore \; \mathcal{C}^S \cap \mathcal{C}^U=\emptyset$.

\label{sec:CC-FG-ZS-SBIR}
\keypoint{Cross-Category Zero-Shot Fine-Grained SBIR.} Unlike category-level, here the aim is to train a \textit{single} model capable of fine-grained instance-level matching from multiple ($N_c$) classes \cite{sain2023clip}. Standard cross-category FG-SBIR datasets (\eg, Sketchy~\cite{sangkloy2016the}) consist of instance-level sketch-photo pairs $\{s_i^j,p_i^j\}_{i=1}^{N_j}{|}_{j=1}^{N_c}$ from $N_c$ categories with $N_j$ sketch-photo pairs per category with {fine-grained association}. Similar to category-level, cross-category FG-SBIR framework trains a backbone feature extractor $\mathcal{B}(\cdot)$ \textit{shared} between sketch and photo branches, over a similar triplet loss (Eq.~\ref{equ:basic_triplet}) but with \textit{hard} triplets, where the negative sample is a \textit{different instance} ($p^j_k;k\neq i$) of the \textit{same class} as the anchor sketch ($s_i^j$) and its matching photo ($p_i^j$)~\cite{sain2023clip}. It further employs an $N_c$-class classification head on the joint embedding space for learning class discrimination~\cite{sain2023clip}.
Following ZS-SBIR paradigm, the classes used for inference are also unseen ($\mathcal{C}^S \cap \mathcal{C}^U=\emptyset$) with the aim of instance-level matching instead of category-level~\cite{sain2023clip}.

While existing works \cite{sain2023clip,yu2016sketch,sain2023exploiting} have designed $\mathcal{B}(\cdot)$ with ImageNet pre-trained VGG-16 \cite{yu2016sketch}, PVT \cite{sain2023exploiting}, ViT \cite{sain2023exploiting}, or foundational models like CLIP~\cite{sain2023clip}, the research question remains as to how can we use a frozen text-to-image diffusion model as a backbone feature extractor $\mathcal{B}(\cdot)$.

\vspace{-0.1cm}
\section{Prompt Learning for SBIR via Frozen SD}
\begin{figure}[t]
    \centering
    \includegraphics[width=\columnwidth]{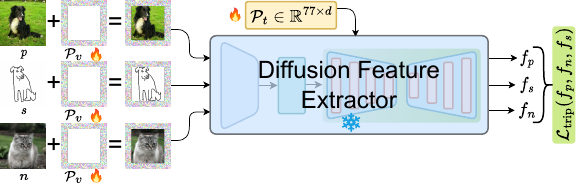}
    \vspace{-0.6cm}
    \caption{Given the frozen SD~\cite{rombach2022high} backbone feature extractor, our method learns a single textual prompt, and sketch/photo-specific visual prompts via triplet loss.}
    \label{fig:arch}
    \vspace{-0.5cm}
\end{figure}

\vspace{-0.2cm}
\keypoint{Overview.} Adapting a frozen SD model~\cite{rombach2022high} as a backbone feature extractor $\mathcal{B}(\cdot)$ is non-trivial. For instance -- \textit{(i)} one off-the-shelf choice is to fine-tune the entire UNet with triplet loss~\cite{yu2016sketch}. However, being pre-trained with a \textit{generative} objective~\cite{rombach2022high}, the UNet will lose its rich semantic knowledge-base when fine-tuned with a \textit{discriminative} loss; \textit{(ii)} learning additional FC layers on top of extracted features via triplet loss (akin to linear probing~\cite{bahng2022exploring}) is also sub-optimal for the same reason; finally, \textit{(iii)} the forward diffusion process~\cite{rombach2022high, ho2020denoising} being stochastic in nature~\cite{rombach2022high} introduces random noise in output feature maps, making them sub-optimal for discriminative learning in the metric space. To tackle these issues related to losing pre-trained knowledge (via fine-tuning or linear probing), we shift to a \textit{prompt learning}~\cite{bahng2022exploring, zhou2022learning}-based approach (\cref{fig:arch}), keeping the weights of the SD model~\cite{rombach2022high} frozen and thus its semantic knowledge-base intact. To further increase the representation-stability of the extracted features, we use \textit{feature ensembling}.

Specifically, we have three salient components: \textit{(i)} a learnable task-specific visual prompt, \textit{(ii)} a learnable continuous textual prompt to harness the visio-linguistic prior of SD~\cite{rombach2022high}, and \textit{(iii)} feature ensembling for additional robustness during inference.

\vspace{-0.2cm}

\vspace{+2mm}
\label{sec:vis_prompt}
\keypoint{Visual Prompt Design.} Originating from NLP literature \cite{shin2020autoprompt}, \textit{prompting} \cite{zhou2022learning, zhou2022conditional} has now become a de facto choice for several vision tasks \cite{zhou2022conditional, bahng2022exploring, jia2022visual}. Prompting involves adapting a large-scale pre-trained model to a specific downstream task without modifying its weights \cite{zhou2022learning, bahng2022exploring}. However, creating task-specific prompts is labour-intensive and requires domain expertise \cite{zhou2022learning, bahng2022exploring}. Prompt learning alleviates this issue by learning a set of context vectors via backpropagation, keeping the model weights frozen \cite{zhou2022learning}. In ViT \cite{dosovitskiy2020an}-based models with vectorised patch-representation, one usually appends additional trainable continuous prompt vectors along with the patch-tokens, for task-specific adaption \cite{jia2022visual}. However, the convolutional input/output (spatial tensor) of SD's UNet denoiser \cite{rombach2022high} prevents direct concatenation of trainable prompt vectors.

Consequently, we opt for visual prompting~\cite{bahng2022exploring} that learns a soft image perturbation in the pixel space to adapt SD model~\cite{rombach2022high} to our problem setup of ZS-SBIR. Given an image $\mathcal{I}\in\mathbb{R}^{h\times w\times 3}$, we add a learnable \textit{spatial} image prompt $\mathcal{P}_v\in\mathbb{R}^{h\times w\times 3}$ whose every ($x,y$) spatial position within the border region (width=$d$) is a trainable continuous $\mathbb{R}^3$-dimensional vector, while the remaining locations are filled with a fixed value of zero and kept frozen. Mathematically,

\vspace{-0.4cm}
\begin{equation}
\scriptsize
\mathcal{P}_v(x,y)=\begin{cases}\mathtt{learnable} & \forall~~~ \left[\begin{matrix} {(x < d)}~\vee~{(x \geq (w-d))}\\{(y < d)}~\vee~{(y \geq (h-d))} \end{matrix}\right]\\0, \mathtt{frozen} & elsewhere\end{cases}
\label{eq:vp}
\end{equation}
\vspace{-0.2cm}

Therefore, the prompt added image $\mathcal{I}_p = \mathcal{I}+\mathcal{P}_v$ is passed through the SD feature backbone to generate the embedded feature as described in \cref{sec:feat_ext}. For a border size $d$, the number of learnable parameters is $2\times c\times d(h+w-2d)$.

\keypoint{Textual Prompt Design.} Text-to-image diffusion models~\cite{rombach2022high, dhariwal2021diffusion} being trained on text-to-image generative objective~\cite{rombach2022high}, works best with \textit{explicit} textual prompts. However, the standard sketch-photo datasets \cite{yu2016sketch, sangkloy2016the} lack associated textual descriptions. Furthermore, generating textual captions via off-the-shelf \textit{image} captioner~\cite{li2022blip} does not fare well for sparse and binary freehand \textit{sketches}~\cite{chowdhury2023scenetrilogy}. To mitigate this, instead of actual textual prompt embedding $\mathbf{T}_p$$=$$\mathcal{T}(\mathbf{p})\in\mathbb{R}^{77\times d}$, we use a learnable continuous textual prompt embedding matrix $\mathcal{P}_t\in\mathbb{R}^{77\times d}$, influencing the SD feature extraction process via cross-attention (\cref{sec:diffusion_background}).

\keypoint{Prompt Training.} As the earlier ($n$$=$$1, 2$) and later ($n$$=$$3, 4$) upsampling layers hold coarse-grained and fine-grained features respectively (\cref{sec:pilot_1}), for category-level ZS-SBIR, we empirically use the mean of the $1^\text{st}$ and $2^\text{nd}$ upsampling layer's global max-pooled features ($t$$=$$273$) $\Hat{f_\mathbf{u}^{1}}\in\mathbb{R}^{1280\times 1}$ and $\Hat{f_\mathbf{u}^{2}}\in\mathbb{R}^{1280\times 1}$, while for cross-category zero-shot FG-SBIR, we concatenate the $3^{rd}$ and $4^\text{th}$ upsampling layer's global max-pooled features ($t$$=$$273$) $\Hat{f_\mathbf{u}^{3}}\in\mathbb{R}^{640\times 1}$ and $\Hat{f_\mathbf{u}^{4}}\in\mathbb{R}^{320\times 1}$ to form a $\mathbb{R}^{960\times 1}$ feature. Now, as the SD feature extraction pipeline is differentiable~\cite{rombach2022high} (although frozen), gradient updates from triplet loss (Eq.\ \ref{equ:basic_triplet}) will be directly backpropagated across UNet and VAE encoder (via reparameterisation-trick~\cite{kingma2013auto}) to update the learnable parts of $\mathcal{P}_v$ (Eq.\ \ref{eq:vp}), and those backpropagated across the UNet will update $\mathcal{P}_t$. In practice, for category-level ZS-SBIR, we use two separate visual prompts for sketch and photo branches, but share a common visual prompt for fine-grained ZS-SBIR. Furthermore, thanks to its implicit semantic knowledge~\cite{rombach2022high}, pre-trained diffusion models can handle multi-category retrieval \textit{without} category-specific fine-tuning~\cite{li2023your}, thus bypassing the need for additional $N_c$-class classification head as used in baseline cross-category zero-shot FG-SBIR (\cref{sec:CC-FG-ZS-SBIR}).

\label{sec:ensemble}
\keypoint{Feature Ensembling during Inference.} Calculating $\mathbf{z}_{t}$ from $\mathbf{z}_{0}$ (during forward diffusion) invokes stochasticity due to the random noise sampling~\cite{rombach2022high}, which could deteriorate the quality of extracted features~\cite{tang2023emergent}. To tackle this, we extract SD features for each image/sketch six times (ablated in \cref{sec:abal}) each from different noise samples, and \textit{ensemble} them by averaging to obtain the final feature.

\vspace{-0.2cm}
\section{Experiments}
\vspace{-0.2cm}
\keypoint{Datasets.\ } We evaluate on three popular sketch-photo datasets -- \textit{(i)} \textbf{Sketchy}~\cite{sangkloy2016the} consists of $12,500$ photos across $125$ classes each having at least $5$ fine-grained paired sketches. While we use the original Sketchy dataset with fine-grained sketch-photo pairing to evaluate cross-category ZS-FG-SBIR, for zero-shot category level setup, we use its extended version~\cite{liu2017deep} with an additional $60,652$ ImageNet photos. Following~\cite{yelamarthi2018zero, sain2023clip}, we use sketches/photos from $104$ and $21$ classes for training and testing respectively. \textit{(ii)} \textbf{Quick, Draw!}~\cite{ha2017neural} has $\sim50M$ sketches categorised into $345$ classes. Following~\cite{dey2019doodle}, we use a subset of it encompassing $330,000$ sketches and $204,000$ photos across $110$ categories split as $80$:$30$ for training:testing. \textit{(iii)} \textbf{TU-Berlin}~\cite{eitz2012humans} houses $204,489$ photos across $250$ categories each having at least $80$ sketches. Following~\cite{dey2019doodle}, we use a training:testing class split of $220$:$30$.

\keypoint{Implementation Details.} We use Stable Diffusion v2.1 \cite{rombach2022high} in all experiments with a CLIP~\cite{radford2021learning} embedding dimension $d=1024$. The visual and textual prompts are trained with a learning rate of $10^{-4}$, keeping the UNet denoiser and the VAE encoder frozen. We learn the prompts for $100$ epochs, on an Nvidia V100 GPU using AdamW~\cite{loshchilov2019decoupled} optimiser with $0.09$ weight decay, and a batch size of $64$.

\keypoint{Evaluation.} For zero-shot category-level setup, following~\cite{sain2023clip, dey2019doodle}, evaluation considers the top $200$ retrieved images to calculate mean average precision (mAP@200) and precision (P@200) on Sketchy while reporting mAP@all and P@100 scores for TU-Berlin. Furthermore, for Quick, Draw!\ dataset we resort to mAP@all and P@200~\cite{sain2023clip}. For cross-category ZS-FG-SBIR evaluation however, we use Acc.@q, which denotes the percentage of sketches with true-matched photos in the top-q retrieved images.

\label{sec:competitors}
\keypoint{Competitors.} We compare against state-of-the-art (SoTA) ZS-SBIR frameworks in two paradigms \textit{viz.}\ ZS-SBIR and ZS-FG-SBIR. Among \textbf{ZS}-SBIR SoTA paradigms, we compare with \textbf{ZS-CVAE}~\cite{yelamarthi2018zero}, \textbf{ZS-CAAE}~\cite{yelamarthi2018zero}, \textbf{ZS-CCGAN}~\cite{dutta2019semantically}, \textbf{ZS-GRL}~\cite{dey2019doodle}, \textbf{ZS-SAKE}~\cite{liu2019semantic}, \textbf{ZS-IIAE}~\cite{hwang2020variational}, \textbf{ZS-Sketch3T}~\cite{sain2022sketch3t}, and \textbf{ZS-LVM}~\cite{sain2023clip}. While the SoTA methods rely on sketch-to-image generation~\cite{yelamarthi2018zero}, word2vec~\cite{mikolov2013distributed} encoding with adversarial learning~\cite{dutta2019semantically}, knowledge-distillation~\cite{liu2019semantic}, test-time adaptation~\cite{sain2022sketch3t}, or pre-trained CLIP models~\cite{sain2023clip}, our method leverages the information-rich internal representation of frozen SD model for generalisation to unseen classes. {Due to the unavailability of diffusion model or visual prompting-based ZS-SBIR frameworks}, we compare against a few self-designed \textbf{B}aselines. \textbf{B-Fine-Tuning} fine-tunes the pre-trained UNet of SD (with a learning rate of $10^{-6}$), while \textbf{B-Linear-Probe} learns an FC layer on top of the SD features, keeping the rest of the model-weights frozen, both via triplet loss~\cite{yu2016sketch}. Next, we extend the baseline triplet loss-based baseline ZS-SBIR with learnable visual prompts (VP) to form \textbf{B-Triplet+VP}. Furthermore, for B-Triplet+VP, we use three different backbone feature extractors \textit{viz.}\ VGG-16~\cite{simonyan2015very}, ResNet50~\cite{he2015delving}, and ViT~\cite{dosovitskiy2020an} pre-trained with discriminative task to compare the contribution of generative pre-training over discriminative. These baselines use the exact same visual prompt dimension (\cref{sec:vis_prompt}) as ours. We extend these triplet-based baselines to the ZS-FG-SBIR setup via \textit{hard} triplet training and an additional classification head (\cref{sec:CC-FG-ZS-SBIR}).

\vspace{-0.2cm}
\subsection{Result Analysis}
\vspace{-0.2cm}
\keypoint{Category Level ZS-SBIR.} Quantitative results shown in \cref{tab:zs-sbir} depict that the naive adaption of pre-trained SD model in B-Fine-Tuning and B-Linear-Probing fails drastically due to loss of generative potential during triplet-based training. {Although B-Triplet+VP (ViT) performs better than B-Triplet+VP (VGG) and B-Triplet+VP (ResNet) on all three datasets, due to its larger backbone, our method supersedes all these baselines with a mAP@200 of $0.746$ on Sketchy, which further verifies our motivation to use the frozen SD model as a backbone feature extractor.} While ZS-SoTA methods offer reasonable performance across all three benchmark datasets owing to the respective zero-shot adaption strategies (\eg, domain translation~\cite{yelamarthi2018zero}, word2vec embedding~\cite{dey2019doodle, dutta2019semantically}, KD~\cite{liu2019semantic}, test-time training~\cite{sain2022sketch3t}, etc.), our method surpasses them with an average mAP@200 gain of $43.96\%$ (Sketchy). ZS-LVM~\cite{sain2023clip} despite being our strongest contender, performs sub-optimally in all benchmarks. The demonstrated efficacy (\cref{tab:zs-sbir}) of frozen SD backbone in category-level ZS-SBIR leads us to the more challenging evaluation setup of ZS-FG-SBIR.

\vspace{-0.1cm}
\begin{table}[!htbp]
    \centering
    \setlength{\tabcolsep}{3.8pt}
    \renewcommand{\arraystretch}{1}
    \notsotiny
    \caption{Results for category-level ZS-SBIR.}
    \vspace{-0.3cm}
    \begin{tabular}{lcccccc}
    \toprule
        \multicolumn{1}{c}{\multirow{2}{*}{Methods}} & \multicolumn{2}{c}{Sketchy~\cite{sangkloy2016the}} & \multicolumn{2}{c}{TU-Berlin~\cite{eitz2012humans}} & \multicolumn{2}{c}{Quick, Draw!~\cite{ha2017neural}} \\\cmidrule(lr){2-3}\cmidrule(lr){4-5}\cmidrule(lr){6-7}
        & mAP@200 & P@200 & mAP@all & P@100 & mAP@all & P@200  \\\cmidrule(lr){1-3}\cmidrule(lr){4-5}\cmidrule(lr){6-7}
        ZS-CAAE~\cite{yelamarthi2018zero}          & 0.156 & 0.260 & 0.005 & 0.003 & --    & --\\
        ZS-CVAE~\cite{yelamarthi2018zero}          & 0.225 & 0.333 & 0.005 & 0.001 & 0.003 & 0.003\\
        ZS-CCGAN~\cite{dutta2019semantically}      & --    & --    & 0.297 & 0.426 & --    & --\\
        ZS-GRL~\cite{dey2019doodle}                & 0.369 & 0.370 & 0.110 & 0.121 & 0.075 & 0.068\\
        ZS-SAKE~\cite{liu2019semantic}             & 0.497 & 0.598 & 0.475 & 0.599 & --    & --\\
        ZS-IIAE~\cite{hwang2020variational}        & 0.373 & 0.485 & 0.412 & 0.503 & --    & --\\
        ZS-Sketch3T~\cite{sain2022sketch3t}        & 0.579 & 0.648 & 0.507 & 0.671 & --    & --\\
        ZS-LVM~\cite{sain2023clip}                 & 0.723 & 0.725 & 0.651 & 0.732 & 0.202 & 0.388\\\cmidrule(lr){1-3}\cmidrule(lr){4-5}\cmidrule(lr){6-7}

        B-Fine-Tuning                              & 0.115 & 0.174 & 0.010 & 0.006 & 0.002 & 0.003\\
        B-Linear-Probing                           & 0.441 & 0.535 & 0.410 & 0.582 & 0.092 & 0.099\\
        B-Triplet+VP (VGG)                         & 0.651 & 0.682 & 0.582 & 0.673 & 0.134 & 0.310\\
        B-Triplet+VP (ResNet)                      & 0.326 & 0.342 & 0.354 & 0.512 & 0.105 & 0.275\\
        B-Triplet+VP (ViT)                         & 0.681 & 0.697 & 0.601 & 0.694 & 0.185 & 0.321\\
        \rowcolor{YellowGreen!40}
        \textbf{\textit{Ours}}                     & \bf0.746 & \bf0.747 & \bf0.680 & \bf0.744 & \bf0.231 & \bf0.397\\
        \bottomrule
    \end{tabular}
    \label{tab:zs-sbir}
    \vspace{-0.1cm}
\end{table}

\keypoint{Cross-Category ZS-FG-SBIR.} Being more challenging than category-level ZS-SBIR, ZS-FG-SBIR is relatively under-explored~\cite{sain2023clip}. Quantitative results in \cref{tab:cc-fg-zs-sbir} show how our method surpasses SoTA ZS-FG-SBIR frameworks \cite{pang2019generalising, sain2023clip, shankar2018generalizing} with an impressive average Acc.@1 margin of $32.49\%$. B-Fine-Tuning and B-Linear-Probing fare poorly here as well confirming our prior observation in the ZS-SBIR setup. Furthermore, our method exceeds all visual prompting-based triplet baselines without the complicacy of learning additional classification head (\cref{sec:competitors}).

\vspace{-0.2cm}
\begin{table}[!htbp]
    \centering
    \setlength{\tabcolsep}{4pt}
    \renewcommand{\arraystretch}{1.1}
    \setlength{\aboverulesep}{0pt}
    \setlength{\belowrulesep}{1pt}
    \notsotiny
    \caption{Results for cross-category ZS-FG-SBIR on Sketchy~\cite{sangkloy2016the}.}
    \vspace{-0.3cm}
    \begin{tabular}{lcc|lcc}
    \toprule
     \multicolumn{1}{c}{\multirow{1}{*}{Methods}}  & Acc.@1 & Acc.@5 & \multicolumn{1}{c}{\multirow{1}{*}{Methods}} & Acc.@1 & Acc.@5\\\cmidrule(lr){1-3}\cmidrule(lr){4-6}

         CC-Gen \cite{pang2019generalising}       & 22.60  & 49.00 &  B-Triplet+VP (VGG)           & 24.20 & 43.61\\
         CC-Grad \cite{shankar2018generalizing}   & 13.40  & 34.90 &  B-Triplet+VP (ResNet)        & 15.61 & 27.64 \\
         CC-LVM \cite{sain2023clip}               & 28.68 & 62.34 &  B-Triplet+VP (ViT)           & 26.11 & 46.81\\\cmidrule(lr){1-3}
         B-Fine-Tuning                            & 1.85 & 6.01   &  \cellcolor{YellowGreen!40}   & \cellcolor{YellowGreen!40} & \cellcolor{YellowGreen!40}\\
         B-Linear-Probing                         & 17.32 & 41.23 &  \multirow{-2}{*}{\cellcolor{YellowGreen!40}\textbf{\textit{Ours}}}   & \multirow{-2}{*}{\cellcolor{YellowGreen!40}\textbf{31.94}} & \multirow{-2}{*}{\cellcolor{YellowGreen!40}\textbf{65.81}}\\ \bottomrule

    \end{tabular}
        \label{tab:cc-fg-zs-sbir}
    \vspace{-0.2cm}
\end{table}

\begin{figure}[!htbp]
    \centering
    \includegraphics[width=1\linewidth]{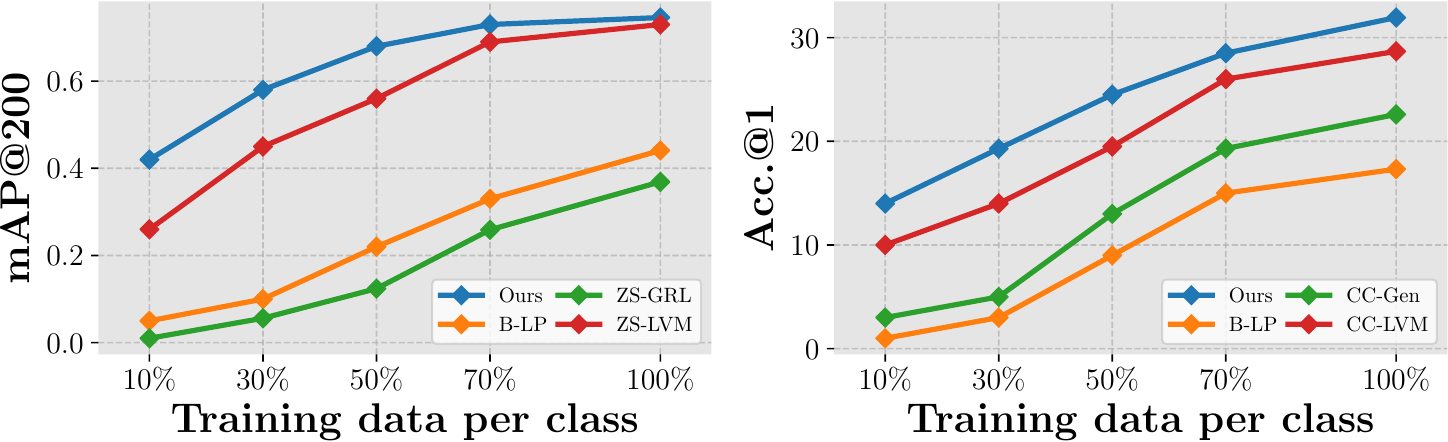}
    \vspace{-0.7cm}
    \caption{Plots showing low-data scenario performance for ZS-SBIR (left) and ZS-FG-SBIR (right) setup on Sketchy~\cite{sangkloy2016the} dataset.}
    \label{fig:low_data}
    \vspace{-0.4cm}
\end{figure}

\keypoint{Performance in Low-Data Scenario.} Existing ZS-(FG)-SBIR models are limited by the availability of paired sketch-photo training data~\cite{sain2023clip, yelamarthi2018zero, dutta2019semantically, dey2019doodle}. To test our method in a \textit{low-data scenario}, we experiment by varying the number of training data ($10\%$, $30\%$, $50\%$, $70\%$, $100\%$) per class from Sketchy~\cite{sangkloy2016the}. Our method remains relatively stable, outperforming (\cref{fig:low_data}) other baseline and SoTAs significantly, preserving almost at par accuracy with \textit{full-scale} training, by harnessing the out-of-distribution generalisation~\cite{ruiz2023dreambooth, tang2023emergent, xu2023open} capability of pre-trained SD~\cite{rombach2022high} model.

\begin{figure*}[!htbp]
    \centering
    \includegraphics[width=1\linewidth]{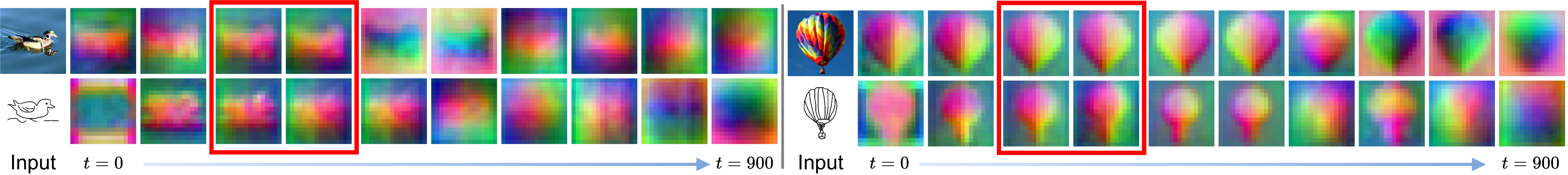}
    \vspace{-0.6cm}
    \caption{PCA representation of SD~\cite{rombach2022high} internal features from $\mathcal{F}_\mathbf{u}^1$ upsampling layers of UNet for different time-steps ($t\in[0,100, ..., 900]$). Different regions of sketch and photo feature maps from $t\in[200,300]$ (highlighted in \red{red}) portray strong semantic feature correspondence (represented by the same colours in the PCA map), while the features from the later time-steps are non-aligned.}
    \label{fig:timestep}
    \vspace{-0.3cm}
\end{figure*}

\subsection{Ablation}
\begin{figure}[!htbp]
    \centering
    \includegraphics[width=1\linewidth]{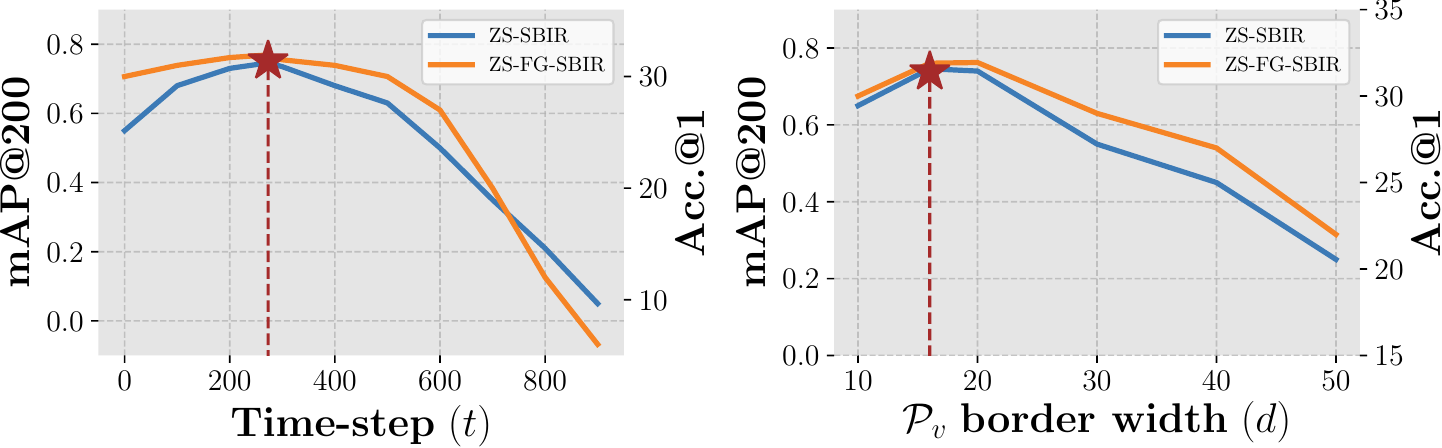}
    \vspace{-0.6cm}
    \caption{Quantitative results on Sketchy~\cite{sangkloy2016the} for ZS-SBIR (mAP@200) and ZS-FG-SBIR (Acc.@1) setup for different denoising time-steps (left) and visual prompt border width (right).}
    \label{fig:abal}
    \vspace{-0.6cm}
\end{figure}

\vspace{-0.2cm}
\label{sec:abal}
\keypoint{[i] Contribution of visual and textual prompting.} To estimate the effectiveness of visual prompts, we train our method without learning the visual prompt, thus eliminating the task-specific adaptation. A significant mAP@200 (Acc.@1) drop of $0.233(14.28)$ on Sketchy (\cref{tab:abal}) in case of \textbf{w/o visual prompt} signifies that adapting SD for ZS-SBIR is incomplete without explicitly learning task-specific prompts in the pixel space. The spatial area of the visual prompt can influence the final performance significantly~\cite{bahng2022exploring}. Varying the visual prompt's border width ($d$) in the range $[10,20,30,40,50]$, we find that $d=16$ produces the optimum mAP@200 (Acc.@1) on Sketchy (\cref{fig:abal} (right)). Secondly, to judge the efficacy of the textual prompts, we replace the learned textual prompts with \textit{null} prompts (\ie, $\mathtt{``~"}$) during inference. The result of \textbf{w/o textual prompt} in \cref{tab:abal} depicts that removing the learned textual prompts further plummets the mAP@200 by $6.83\%$.

\keypoint{[ii] Importance of feature ensembling.} Feature ensembling helps reduce the effect of stochastic noising (during forward diffusion) of latent images in the final feature maps. Removing it further destabilises the framework causing an additional mAP@200 drop of $0.021$ on Sketchy (\cref{tab:abal}). We posit that this drop is due to the absence of the additional smoothing regularisation provided by feature ensembling that reinforces discriminative learning in the metric space. Experimenting with different sizes ($[1,8]$), we found that feature ensembling six SD features gives optimum accuracy across all datasets.

\keypoint{[iii] Do $\mathcal{F}_{\mathbf{d}}^{n}$ layers hold representative features?} We also evaluate the ZS-SBIR and ZS-FG-SBIR performance by using UNet downsampling layer $\mathcal{F}_{\mathbf{d}}^{n}$ features. We train our model with features from $\{\mathcal{F}_{\mathbf{d}}^{n}\}_{n=1}^2$ and $\{\mathcal{F}_{\mathbf{d}}^{n}\}_{n=3}^4$ for ZS-FG-SBIR and ZS-SBIR setup respectively. Nonetheless, sub-optimal mAP@200 and Acc.@1 on Sketchy in case of \textbf{$\mathcal{F}_{\mathbf{d}}^{n}$ features} (\cref{tab:abal}) render the UNet downsampling layer features unsuitable for the tasks of ZS-(FG)-SBIR.

\keypoint{[iv] Which feature-level of $\mathcal{F}_{\mathbf{u}}^{n}$ is optimum?} As evident from \cref{fig:teaser}, different resolution-levels of $\mathcal{F}_{\mathbf{u}}^{n}$ contain features of different \textit{semantic-granularity}. To select the optimal set of layers for feature extraction, we performed a $2D$-grid search separately for ZS-SBIR and ZS-FG-SBIR setups on Sketchy \cite{sangkloy2016sketchy}. Results shown in \cref{tab:idx} unfold that a combination of $n=1,2$ and $n=3,4$ works best for ZS-SBIR and ZS-FG-SBIR respectively.

\keypoint{[v] Do all time-steps yield useful features?} As seen in \cref{fig:timestep}, the cross-modal feature maps are most \textit{semantically-aligned} (\textit{same} colour) around $t\in[200,300]$. Calculating the average mAP@200 (Acc.@1) on Sketchy over varying $t$ shows $t=273$ to generate optimal features for both ZS and ZS-FG-SBIR. As seen in \cref{fig:abal} (left), our method is quite robust to the selection of $t$, as a wide range of $t$ produces mAP@200 (Acc.@1) well over the baselines.

\begin{table}[!htbp]
\parbox{.5\columnwidth}{
    \setlength{\tabcolsep}{3pt}
    \renewcommand{\arraystretch}{1.2}
    \centering
    \notsotiny
    \caption{Ablation on design.}
    \vspace{-0.3cm}
    \label{tab:abal}
    \begin{tabular}{lcc}
    \toprule
    \multicolumn{1}{c}{\multirow{2}{*}{Methods}} & \multicolumn{2}{c}{Sketchy~\cite{sangkloy2016the}}\\
    \cmidrule(lr){2-3}
                                             & mAP@200   & Acc.@1\\
    \cmidrule(lr){1-3}
    w/o visual prompt                        & 0.513     & 17.66\\
    w/o textual prompt                       & 0.695     & 22.71\\
    w/o feature ensemble                     & 0.725     & 29.47\\
    $\mathcal{F}_{\mathbf{d}}^{n}$ features  & 0.365     & 18.22\\
    \rowcolor{YellowGreen!40}
    \textbf{\textit{Ours-full}}              & \bf0.746  & \bf31.94\\
    \textit{\graytext{Avg. Improvement}}     & \graytext{\textit{0.171}}  & \graytext{\textit{9.92}}\\
    \bottomrule
    \end{tabular}
}\hfill
\parbox{.45\columnwidth}{
    \centering
    \setlength{\tabcolsep}{3pt}
    \renewcommand{\arraystretch}{1.07}
    \notsotiny
    \caption{Choice of $\mathcal{F}_{\mathbf{u}}^{n}$ index.}
    \vspace{-0.3cm}
    \label{tab:idx}
    \begin{tabular}{cccc|cc}
         \toprule
         \multicolumn{4}{c|}{$n$} & \multicolumn{2}{c}{Sketchy~\cite{sangkloy2016the}}\\\cmidrule(lr){1-4}\cmidrule(lr){5-6}
         $1$         & $2$         & $3$         & $4$         & mAP@200 & Acc.@1\\\cmidrule(lr){1-6}
         \greencheck & \redcross & \redcross & \redcross       & 0.431      & 10.40\\
         \greencheck & \greencheck & \redcross & \redcross     & \bf0.746   & 15.21\\
         \greencheck & \greencheck & \greencheck & \redcross   & 0.712      & 25.44\\
         \greencheck & \greencheck & \greencheck & \greencheck & 0.613      & 29.36\\
         \redcross & \redcross & \redcross & \greencheck       & 0.287      & 20.61\\
         \redcross & \redcross & \greencheck & \greencheck     & 0.352      & \bf31.94\\
         \redcross & \greencheck & \greencheck & \greencheck   & 0.408      & 30.21\\\bottomrule
    \end{tabular}
}
\vspace{-0.10cm}
\end{table}

\vspace{-0.2cm}

\vspace{0.25cm}
\section{Extension to Sketch+Text-based Retrieval}
Stable Diffusion~\cite{rombach2022high} is trained with large-scale image-text pairs with a generative objective, where the model learns the \textit{visio-linguistic interactions} between image and text via cross-attention~\cite{rombach2022high}. Presuming the non-availability of textual captions/class labels for standard SBIR tasks, we used learnable textual prompts in our proposed method. However, here, we extend our method to \textit{Zero-Shot Sketch+Text-Based Image Retrieval} (ZS-STBIR) where we aim to improve the extracted feature-quality with readily available textual captions/class labels. In other words, While most of the existing SBIR backbones (\eg, VGG-16~\cite{simonyan2015very}, PVT~\cite{wang2021pyramid}, CLIP~\cite{radford2021learning}, etc.) are a function of image \textit{only} -- $\mathcal{B}(\mathcal{I})$, the proposed stable-diffusion feature extractor could be used to extract joint visio-linguistic features where the visual ($\mathcal{I}$) and textual ($\mathbf{p}$) modality interacts with each other and outputs a \textit{text-enhanced} feature implicitly as {$\mathcal{F}_\theta(\mathcal{I}, \mathbf{p})$}. Notably, although CLIP training involves contrastive learning~\cite{radford2021learning} using image-text pairs, the visual and textual encoders being \textit{independent} of each other, \textit{do not} directly influence the feature extraction process during inference. While several existing works~\cite{sangkloy2022sketch, chowdhury2023scenetrilogy, song2017fine} attempt to address this joint sketch+text multi-modal feature learning either via complicated invertible neural network~\cite{chowdhury2023scenetrilogy}, CLIP-fine-tuning~\cite{sangkloy2022sketch}, or allegedly unstable quadruplet loss~\cite{song2017fine}, we simply pass the associated textual captions in $\mathcal{F}_\theta$ in place of the learnable textual prompts.

In practice, for category-level ZS-SBIR and cross-category ZS-FG-SBIR, we additionally feed the class-name (thus \textit{sketch+text}) in the form of a fixed handcrafted prompt~\cite{zhou2022learning} $\mathtt{``a~photo~of~[CLASS]"}$. Whereas, for scene-level retrieval~\cite{chowdhury2023scenetrilogy}, we use the readily available paired captions from standard scene-level SBIR datasets~\cite{chowdhury2022fs, gao2020sketchycoco}. Overall, a simple off-the-shelf extension of our SD-based feature extractor depicts competitive or better STBIR performance on several benchmark datasets (\cref{tab:cat_adapt}).

\vspace{-0.3cm}
\begin{table}[!htbp]
\setlength{\tabcolsep}{1pt}
\renewcommand{\arraystretch}{1.1}
\centering
\caption{Results for sketch+text-based image retrieval.}
\vspace{-0.3cm}
\label{tab:cat_adapt}
\notsotiny
\begin{tabular}{p{15mm}cccccccc}
\toprule
\multicolumn{1}{c}{\multirow{3}{*}{Methods}} & \multicolumn{2}{c}{Category-level} & \multicolumn{2}{c}{Fine-grained} & \multicolumn{4}{c}{Scene-level} \\\cmidrule(lr){2-3}\cmidrule(lr){4-5}\cmidrule(lr){6-9}
 & \multicolumn{2}{c}{Sketchy~\cite{sangkloy2016the}} & \multicolumn{2}{c}{Sketchy~\cite{sangkloy2016the}} & \multicolumn{2}{c}{FS-COCO~\cite{chowdhury2022fs}} & \multicolumn{2}{c}{Sketchy-COCO~\cite{gao2020sketchycoco}} \\\cmidrule(lr){2-3}\cmidrule(lr){4-5}\cmidrule(lr){6-7}\cmidrule(lr){8-9}
& mAP@200 & P@200 & Acc.@1 & Acc.@5 & Acc.@1 & Acc.@10 & Acc.@1 & Acc.@10 \\
\cmidrule(lr){1-3}\cmidrule(lr){4-5}\cmidrule(lr){6-7}\cmidrule(lr){8-9}

QST~\cite{song2017fine}                       & 0.204 & 0.213 & 8.36 & 16.17 & 25.1 & 54.5 & 38.9 & 87.9\\
TaskFormer~\cite{sangkloy2022sketch}          & 0.491 & 0.501 & 21.02 & 52.34 & 24.9 & 55.1 & 39.0 & 88.2\\
SceneTrilogy~\cite{chowdhury2023scenetrilogy} & 0.601 & 0.672 & 27.52 & 60.13 & 25.7 & 55.2 & 39.5 & 88.7\\\cmidrule(lr){1-3}\cmidrule(lr){4-5}\cmidrule(lr){6-7}\cmidrule(lr){8-9}

B-Linear-Probing                              & 0.562 & 0.613 & 19.87 & 44.21 & 14.8 & 43.7 & 29.7 & 76.2\\
\rowcolor{YellowGreen!40}
\textit{\textbf{Ours}}                        &\bf0.751  &\bf0.762 & \bf33.31 &\bf68.44  &\bf26.3 & \bf56.6 & \bf40.8 & \bf89.6\\
\bottomrule
\end{tabular}
\vspace{-0.3cm}
\end{table}

\vspace{-0.2cm}
\section{Conclusion}
\vspace{-0.2cm}
This paper for the first time proposes a novel pipeline to adapt the frozen Stable Diffusion as a backbone feature extractor for both \textit{category-level} and \textit{cross-category fine-grained ZS-SBIR} tasks. With clever usage of visual and textual prompting, our method adapts the pre-trained model to the task at hand without further fine-tuning. Extensive experimental results on several benchmark datasets depict that the proposed method outperforms state-of-the-art ZS-SBIR methods. Furthermore, we perform thorough analytical experiments to establish the best practices for leveraging frozen stable diffusion models as a ZS-SBIR backbone. Lastly, harnessing the inherent visio-linguistic capability of stable diffusion, we extend our pipeline to \textit{sketch+text}-based SBIR enabling the practical sketch+text-based retrieval at category, fine-grained and scene-level scenarios.

{
    \small
    \bibliographystyle{ieeenat_fullname}
    \bibliography{arxiv}
}

\clearpage

\twocolumn[{\centering{\Large \textbf{Supplementary material for\\ Text-to-Image Diffusion Models are Great Sketch-Photo Matchmakers}\par}\vspace{0.3cm}
	{\MYhref[cvprblue]{https://subhadeepkoley.github.io}{Subhadeep Koley}\textsuperscript{1,2} \hspace{.2cm} \MYhref[cvprblue]{https://ayankumarbhunia.github.io}{Ayan Kumar Bhunia}\textsuperscript{1} \hspace{.2cm} \MYhref[cvprblue]{https://aneeshan95.github.io}{Aneeshan Sain}\textsuperscript{1} \hspace{.2cm}  \MYhref[cvprblue]{https://www.pinakinathc.me}{Pinaki Nath Chowdhury}\textsuperscript{1}\\ \MYhref[cvprblue]{https://www.surrey.ac.uk/people/tao-xiang}{Tao Xiang}\textsuperscript{1,2} \hspace{.2cm} \MYhref[cvprblue]{https://www.surrey.ac.uk/people/yi-zhe-song}{Yi-Zhe Song}\textsuperscript{1,2} \\
\textsuperscript{1}SketchX, CVSSP, University of Surrey, United Kingdom.  \\
\textsuperscript{2}iFlyTek-Surrey Joint Research Centre on Artificial Intelligence.\vspace{0.1cm}\\
{\tt\small \{s.koley, a.bhunia, a.sain, p.chowdhury, t.xiang, y.song\}@surrey.ac.uk}\par\vspace{0.5cm}}}
]

\section*{{A. Time and computational complexity}} Unlike the \textit{iterative inference} of text-to-image generation via stable diffusion (SD) model~\cite{rombach2022high}, diffusion-based feature extraction needs a \textit{single-step inference} (\cref{sec:feat_ext}). {Most importantly, instead of running SD model six times, we resort to an \textit{efficient implementation} where we repeat the query sketch tensor six times along batch dimension and use a set of different random noises to extract six distinct SD features \textit{simultaneously} in \textit{one} step. Thus, the complexity and runtime \textit{do not} scale \textit{linearly}.} Instead, it takes a similar running time compared to competing SoTAs for the same input size. For instance, our diffusion feature extraction (with ensembling) takes $0.85$ms \vs ZS-LVM's \cite{sain2023clip} $0.83$ms or B-Triplet+VP (VGG)'s $76$ms for a $224\times 224$ image on a single Nvidia V100 GPU. Performing feature ensembling to boost performance and stability (\cref{sec:ensemble}) would increase the inference time slightly ($0.82
\rightarrow 0.85$ms). However, in case of a computation bottleneck, one may avoid this with a slight dip in performance (\eg, \textit{Sketchy}: mAP@200 $0.746 \rightarrow 0.725$;  \textit{TU-Berlin}: mAP@all $0.680 \rightarrow 0.671$; \textit{Quick, Draw!}: mAP@all $0.231 \rightarrow 0.220$). Notably, even without feature ensembling, our method surpasses the next best method (\ie ZS-LVM \cite{sain2023clip}) on \textit{all} $3$ benchmark datasets. Consequently, we leave the choice of utilising this gain provided by ensembling (at a slight cost of inference time) to the end-users.

\section*{{B. Performance-complexity trade-off}} Even \textit{with} feature ensembling, our method takes $0.85$ms to extract a query-sketch feature (for a $224\times 224$ sketch) compared to $0.83$ms of our closest competitor (ZS-LVM \cite{sain2023clip}), which is only {$\tiny \sim$}$2.4\%$ higher, yet boosts Acc.@1 by $11.4\%$ (ZS-FG-SBIR on Sketchy). While ZS-LVM \cite{sain2023clip} takes $9.46$G FLOPs (CLIP-ViT-B/32) to process a sketch of size $224\times 224$, our method uses $1.29$G FLOPs, which is $7.33\times$ lower, while boosting mAP@all by $14.4\%$ on the Quick, Draw! dataset.

\section*{{C. Ablating Stable Diffusion versions}} We ablate multiple SD \cite{rombach2022high} versions on Sketchy \cite{sangkloy2016the} dataset in \cref{tab:sd_abal}. While SD v1.x models utilise CLIP \cite{radford2021learning} text encoder during their pre-training, v2.x models resort to much larger-scale OpenCLIP \cite{cherti2023reproducible}. Evidently, SD v2.x models perform better than v1.x ones with v2.1 achieving the highest score. This is likely due to v2.x models' adaptation of the much larger-scale OpenCLIP \cite{cherti2023reproducible} encoder during pre-training.

\begin{table}[!htbp]
    \setlength{\tabcolsep}{10pt}
    \centering
    \small
    \caption{Ablating SD versions.}
    \vspace{-0.2cm}
    \label{tab:sd_abal}
    \begin{tabular}{ccc}
    \toprule
    \multicolumn{1}{c}{\multirow{2}{*}{SD version}} & \multicolumn{2}{c}{Sketchy \cite{sangkloy2016the}}\\
    \cmidrule(lr){2-3}
     & mAP@200 & Acc.@1\\
    \cmidrule(lr){1-3}
    v1.4                                       & 0.726  & 28.93\\
    v1.5                                       & 0.730  & 29.81\\ \cmidrule(lr){1-3}
    v2.0                                       & 0.738  & 30.21\\
    \rowcolor{YellowGreen!40}
    \textbf{{v2.1 (Ours)}}                     & \bf0.746  & \bf31.94\\
    \bottomrule
    \end{tabular}
    \vspace{-3.5mm}
\end{table}

\section*{{D. Result across different ensemble sizes}} 
\begin{figure}[!htbp]
    \centering
    \includegraphics[width=\linewidth]{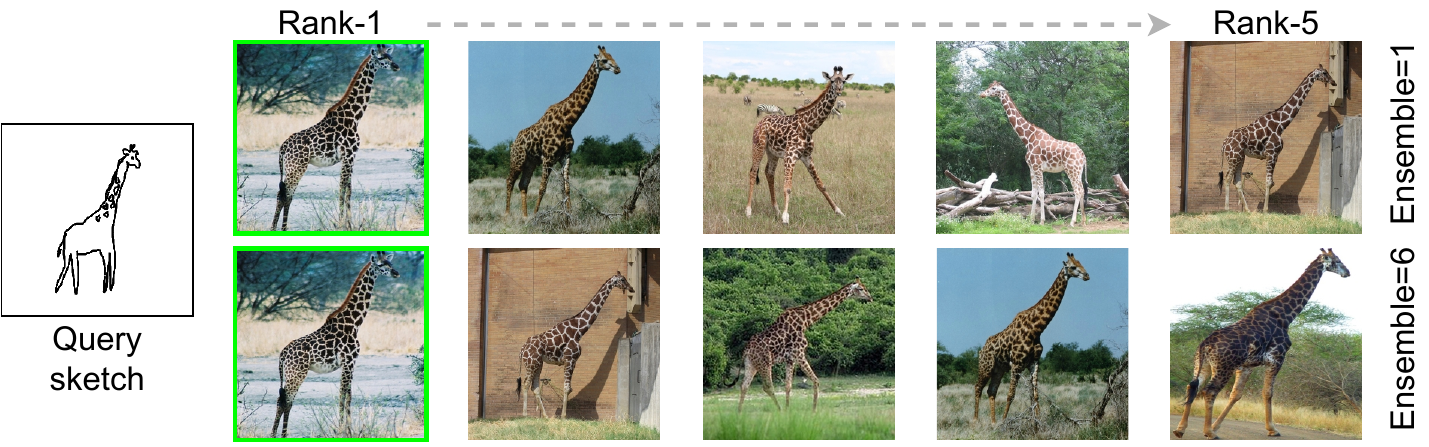}
    \vspace{-0.6cm}
    \caption{qualitative results for different ensemble sizes.}
    \label{fig:qual_1}
\end{figure}

\cref{fig:qual_1} depicts qualitative results for ZS-FG-SBIR on Sketchy across different runs with different ensemble sizes.

\vspace{-0.3cm}

\end{document}